\documentclass[10pt,twocolumn,letterpaper]{article}

\usepackage[pagenumbers]{iccv} 

\definecolor{iccvblue}{rgb}{0.21,0.49,0.74}
\usepackage[pagebackref,breaklinks,colorlinks,allcolors=iccvblue]{hyperref}
\definecolor{b1}{RGB}{223,101,97}
\definecolor{b2}{RGB}{233,161,162}
\definecolor{b3}{RGB}{250,232,209}
\usepackage{colortbl}
\usepackage{xcolor}
\usepackage{multirow}
\usepackage{bm}
\usepackage{algorithm,algorithmic}
\usepackage{url}
\usepackage{graphicx}
\usepackage{amsmath}
\usepackage{accents}
\usepackage{amsthm}
\usepackage{float}
\usepackage{multicol}
\usepackage[belowskip=-10pt,aboveskip=0pt]{caption}

\def\paperID{} 
\def\confName{}
\def\confYear{}

\makeatletter
\def\widebar{\accentset{{\cc@style\underline{\mskip14mu}}}}
\makeatother

\title{EVA-Gaussian: 3D Gaussian-based Real-time Human Novel View Synthesis under Diverse Multi-view Camera Settings}

\author{Yingdong Hu\textsuperscript{\rm ~~1}~~~Zhening Liu\textsuperscript{\rm ~~1}~~~Jiawei Shao\textsuperscript{\rm 2}~~~Zehong Lin\textsuperscript{\rm 1}$^*$ \ Jun Zhang\textsuperscript{\rm 1}\\
\textsuperscript{\rm 1} The Hong Kong University of Science and Technology \\
\hspace{0.3cm} \textsuperscript{\rm 2} Institute of Artificial Intelligence (TeleAI), China Telecom
}

\begin{document}

\twocolumn[{%
\maketitle
\vspace{-28pt}
\begin{figure}[H]
    \centering
    \hsize=\textwidth %
    \includegraphics[width=0.86\textwidth]{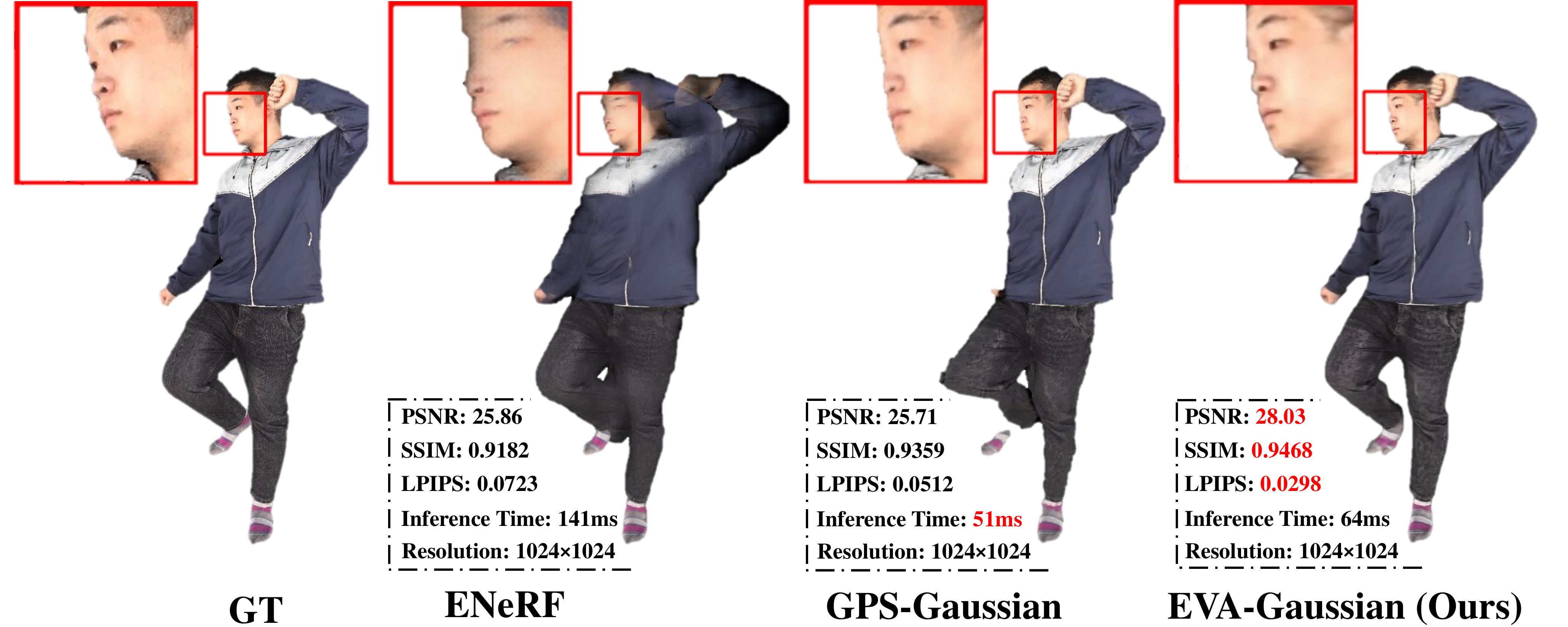}
    \caption{Qualitative comparison of novel view synthesis on the THuman2.0 dataset, with the angle between the stereo views being 72 degree and GT representing the ground truth. We compare our proposed EVA-Gaussian against the state-of-the-art approaches GPS-Gaussian \citep{gpsgs} and ENeRF \citep{enerf}. The quantitative metrics of PSNR$\uparrow$, SSIM$\uparrow$, LPIPS$\downarrow$, and inference time$\downarrow$ demonstrate that EVA-Gaussian achieves superior reconstruction quality, while enabling real-time reconstruction under sparse-view conditions and high-resolution settings.}
    \label{fig:first_visual}
    \vspace{12pt}
\end{figure}
}]

\def\thefootnote{*}\footnotetext{Corresponding author, eezhlin@ust.hk.} 

\begin{abstract}
Feed-forward based 3D Gaussian Splatting methods have demonstrated exceptional capability in real-time novel view synthesis for human models. However, current approaches are confined to either dense viewpoint configurations or restricted image resolutions. These limitations hinder their flexibility in free-viewpoint rendering across a wide range of camera view angle discrepancies, and also restrict their ability to recover fine-grained human details in real time using commonly available GPUs.
To address these challenges, we propose a novel pipeline named \emph{EVA-Gaussian} for 3D human novel view synthesis across diverse multi-view camera settings. Specifically, we first design an Efficient Cross-View Attention (EVA) module to effectively fuse cross-view information under high resolution inputs and sparse view settings, while minimizing temporal and computational overhead. Additionally, we introduce a feature refinement mechianism to predict the attributes of the 3D Gaussians and assign a feature value to each Gaussian, enabling the correction of artifacts caused by geometric inaccuracies in position estimation and enhancing overall visual fidelity.
Experimental results on the THuman2.0 and THumansit datasets showcase the superiority of EVA-Gaussian in rendering quality across diverse camera settings. 
\end{abstract}
\vspace{-15pt}

\section{Introduction}

3D reconstruction and novel view synthesis have long been fundamental yet complex tasks in visual data representation and computer vision. 
Recent advancements in fast 3D reconstruction and novel view synthesis for humans have shown immense potential in applications such as holographic communication, real-time teaching, and augmented/virtual reality (AR/VR), where time efficiency and the clarity of human body representations are critical for delivering satisfactory user experiences. Nonetheless, existing methods either rely on dense input views and precise templates as prior knowledge \citep{3dgs-avatar,gart,gaussianavatar,gomavatar,hugs,generalizableHG} or are restricted to specific camera poses \citep{gpsgs, tele-aloha}. None of these approaches has fully developed a pipeline for real-time human reconstruction under diverse, especially sparse, camera viewpoints while maintaining high-resolution input quality.

In recent years, Neural Radiance Fields (NeRFs) \citep{nerf} have emerged as a promising technique for 3D reconstruction. These models employ neural networks to predict the color and density of sampled 3D points along camera rays and aggregate these predictions to synthesize novel images with high fidelity. 
Despite their effectiveness, NeRFs suffer from substantial time consumption during both the training and rendering phases. Although various advancements, such as multi-resolution hash encoding \citep{instantngp} and feed-forward neural scene prediction \citep{pixelnerf, murf}, have been achieved to mitigate these inefficiencies
, the achievable speeds remain insufficient for real-time applications.

More recently, 3D Gaussian Splatting (3DGS) \citep{3dgs} has been introduced as a solution to the rendering bottleneck. 3DGS utilizes a set of discrete Gaussian representations to model complex 3D scenes and leverages the $\alpha$-blending technique to enable real-time novel view synthesis.
In the field of 3D human avatar reconstruction, previous works \citep{3dgs-avatar,gart,gaussianavatar,gomavatar,hugs,generalizableHG} have employed 3DGS as a representation for humans to achieve animatable full-body human avatar reconstruction. These methods, however, rely on precise human templates as priors, and their training and reconstruction processes can span from minutes to hours, which impedes their use in real-time applications such as holographic communication.
While a feed-forward human reconstruction method \citep{gpsgs} has achieved fast reconstruction and real-time rendering with stereo inputs at a resolution of 1024, its reliance on stereo settings and limited camera angle variations restricts its reconstruction quality under sparse camera settings and lead to sub-optimal performance when more than two input views are utilized. 
 
To address these limitations and enable high-resolution real-time 3D reconstruction of humans across diverse camera positions and varying numbers of cameras, we propose \emph{EVA-Gaussian}, a novel 3D Gaussian-based pipeline for real-time human novel view synthesis. Our method attaches 3D Gaussians to the surface of the human body through multi-view depth estimation and aligns their positions closely with point cloud locations. A key innovation is the introduction of an Efficient cross-View Attention (EVA) module for multi-view 3D Gaussian position estimation (see Sec. \ref{Gaussian Position Estimation}). Specifically, we employ a U-Net \citep{unet} as the backbone and incorperate a dedicated window-embedded cross-view attention mechanism to infer multi-view position correspondences. This design leverages the observation that, in most human capture systems, human subjects are typically centered within each frame. As a result, the epipolar lines tend to align nearly parallel to the x-axis of the image, which enables us to significantly reduce the memory for multi-view attention while maintaining accuracy under large view discrepancies.
Besides, we incorporate a Gaussian attribute estimation module that takes the EVA output and the original RGB images as input to estimate the remaining 3D Gaussian attributes (see Sec. \ref{Gaussian Attributes Estimation}). Furthermore, we embed an additional attribute, referred to as feature, into each Gaussian for further feature splatting and image quality refinement, thereby mitigating the position estimation errors introduced by the EVA module (see Sec. \ref{section Feature Splatting and Refinement}). In addition, we introduce an anchor loss to regularize the opacity and scale attributes of the 3D Gaussians (see Sec. \ref{Regularization}). This ensures consistency between the point cloud depth and the 3D Gaussian position map, enhancing overall stability and accuracy.
We conduct extensive experiments on the THuman2.0 \citep{thuman2.0} and THumanSit \citep{thumansit} datasets. The results, as exemplified in Fig. \ref{fig:first_visual}, demonstrate that EVA-Gaussian outperforms existing feed-forward synthesis approaches in rendering quality, while enabling real-time reconstruction and rendering. Moreover, our approach generalizes well to settings with varying numbers of cameras and significant changes in camera viewpoint angles. Our main contributions are summarized as follows:
\begin{itemize}
    \item We propose \emph{EVA-Gaussian}, a novel pipeline for fast feed-forward 3D human reconstruction that comprises three main stages: 1) 3D Gaussian position estimation, 2) 3D Gaussian attributes estimation, and 3) feature refinement.
    
    \item We introduce an EVA module to enhance multi-view correspondence retrieval, leading to improved 3D Gaussian position estimation and enhanced novel view synthesis under diverse view numbers and sparse camera settings.
    
    \item We embed a feature value to each 3D Gaussian through a feed-forward neural network and employ a recurrent feature refiner that fuses splatted images and feature maps to mitigate artifacts caused by position estimation errors. 
    
    \item Extensive experiments on the THuman2.0 and THumansit datasets demonstrate the effectiveness and superiority of our proposed pipeline over existing methods in terms of rendered novel view quality and inference speed, especially under sparse camera settings.
\end{itemize}

\section{Related Works}

\noindent\textbf{Instant Human Reconstruction.} Recovering a 3D human model from sparse view inputs and generating novel views in real time has been a long-standing challenge in computer vision and graphics. Despite continuous advancements, achieving both accuracy and efficiency remains an unresolved problem. PIFu \citep{pifu} is one of the first methods to successfully reconstruct the human surface and its color map from one or several RGB images, leveraging the strong spatial encoding capabilities of implicit function representations. However, it remains constrained by low speed and resolution. Subsequent works attempt to address these limitations by employing specific rendering methods \citep{realtimepifu,pifuhd} or incorperating additional depth information \citep{thuman2.0}. Nonetheless, their reliance on traditional 3D representations results in suboptimal rendering quality and efficiency.

\noindent\textbf{3DGS-based Human Avatar Reconstruction.}
3D Gaussian Splatting has recently emerged as an effective technique for 3D human reconstruction.
However, most previous works \citep{gart, gaussianavatar, gomavatar, 3dgs-avatar, hugs, humansplat} bind 3D Gaussians to a predefined human mesh model, such as SMPL \citep{smpl} or SMPL-X \citep{smplx}. This approach generates 3D Gaussians and human models in a canonical space and then transforms them to match the target human pose using predefined weights.
This iterative binding process, however, is extremely time-consuming. Moreover, these methods require human templates as inputs at each frame, which incurs extra computational costs and potentially misleads the reconstruction due to errors in pose estimation. These limitations significantly hinder their applicability in real-world scenarios.

\noindent\textbf{Fast Generalizable 3D Reconstruction.} In the field of NeRF rendering, pixelNeRF \citep{pixelnerf} pioneers the approach of predicting per-pixel features from a single image in a feed-forward manner for 3D reconstruction. While subsequent works \citep{mvsnerf, ibrnet, enerf} have followed this feed-forward NeRF pipeline, they still suffer from the extensive time consumption of the NeRF rendering process. Besides, their reconstruction results are often unsatisfactory under sparse camera settings. The introduction of 3DGS has helped mitigate the rendering speed issue of high-quality novel view synthesis. Notably, pixelSplat \citep{pixelsplat} and Splatter Image \citep{splatterimage} are the first to combine feed-forward inference with 3DGS, which predict 3D Gaussian attributes for each pixel and project them back into the 3D space for real-time novel view synthesis. Nevertheless, they still struggle with inaccurate estimation of Gaussian positions. MVSplat \citep{mvsplat} and MVGaussian \citep{FGGS} address this issue by leveraging cross-attention mechanisms and cost-volume modules, thereby achieving superior novel view quality. Moreover, latentSplat \citep{latentsplat} attaches a latent vector to each 3D Gaussian and refines novel views through a diffusion decoder and generative loss, significantly improving image quality in extrapolation views. Despite these advancements, existing methods fail to fully exploit prior knowledge about human images and camera settings, which limits their performance on real-time human reconstruction and novel view synthesis.


The work most closely related to ours is GPS-Gaussian \citep{gpsgs}, which proposes a stereo matching network for 3D Gaussian position estimation and employs two 3-layer U-Nets to predict 3D Gaussian scales, rotations, and opacities. Although GPS-Gaussian demonstrates the potential for real-time human reconstruction and novel view synthesis, it suffers from severe distortions under sparse camera settings and mismatch across multiple viewpoints. Subsequent works have attempted to alleviate these issues. For instance, Tele-Aloha \citep{tele-aloha} introduces an image blending and cascaded disparity estimation method for human reconstruction with four input views. However, this approach is tailored to a specific system and struggles to generalize to sparser camera settings. Although GHG \citep{generalizableHG} achieves real-time 3D Gaussian-based human novel view synthesis in a feed-forward manner, it requires additional human template priors, thus inheriting the limitations of template-based methods. In contrast, our method eliminates the need for human templates and is specifically designed to generalize effectively across diverse sparse camera settings.

\begin{figure*}[t]
    \centering
    \includegraphics[width=0.82\textwidth]{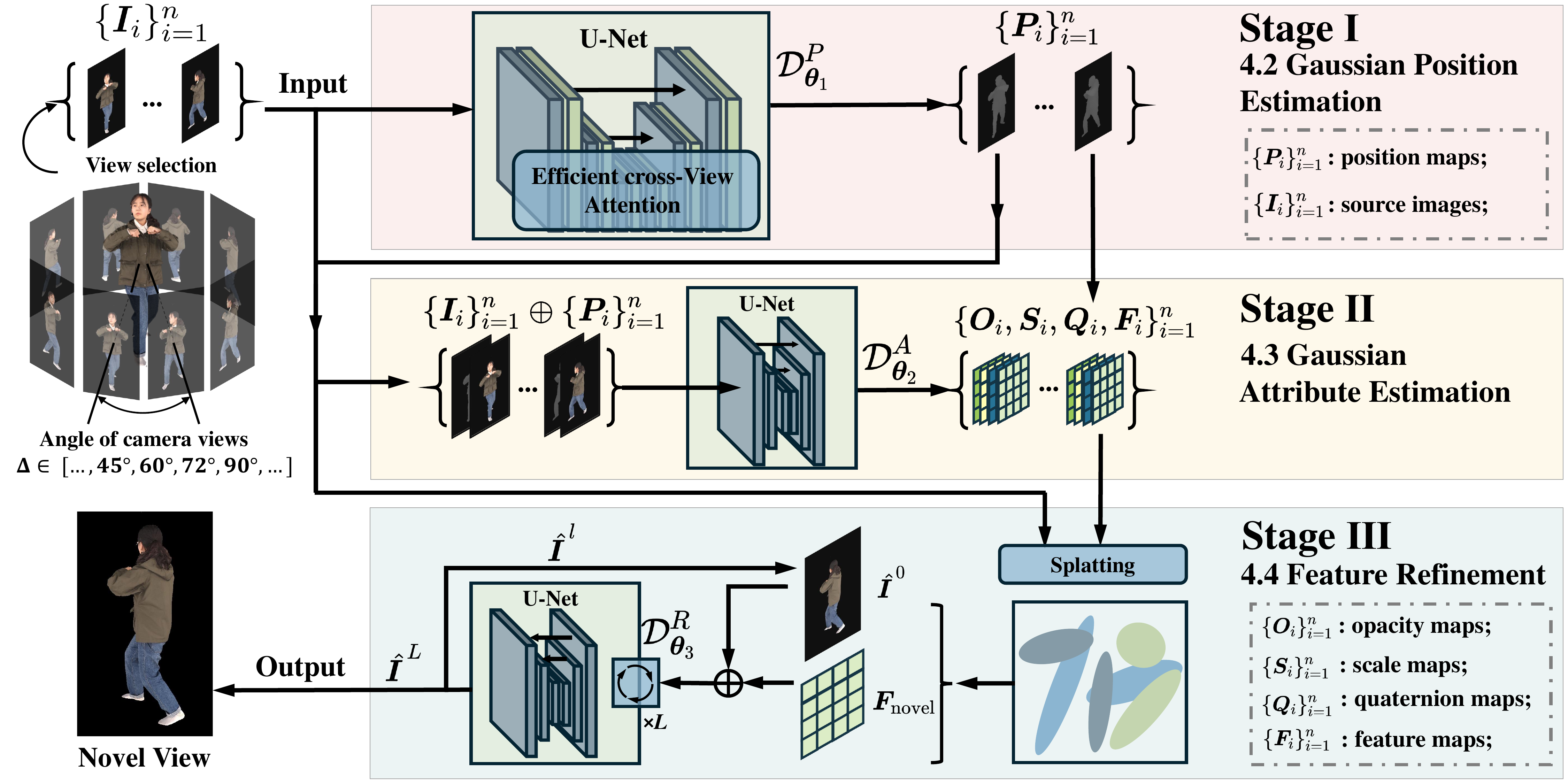}
    \caption{\textbf{Framework of EVA-Gaussian}. EVA-Gaussian takes sparse-view images captured around a human subject as input and performs three key stages: (1) estimating the positions of 3D Gaussians, (2) inferring the remaining attributes (i.e., opacities, scales, quaternions, and features) of these Gaussians, and (3) refining the output image in a recurrent manner.}
    \label{fig:network_overview}
\vspace{-8pt}
\end{figure*}

\section{Methodology}
\subsection{Overview}
\label{section Overview}
In this paper, we focus on fast human 3D reconstruction and novel view synthesis under diverse camera settings. Our objective is to reconstruct a 3D scene from a set of $n$ sparse-view RGB images $\{\bm{I}_i\}_{i=1}^{n}, \bm{I}_i \in \mathbb{R}^{H\times W\times3}$, captured from different viewpoints surrounding a human subject, where the angle between any two adjacent camera views is denoted by $\Delta$, and synthesize arbitrary novel view images at any camera position in real time. To achieve this, we propose \emph{EVA-Gaussian}, a method that utilizes deep neural networks and 3D Gaussian Splatting to enhance novel image quality while achieving real-time reconstruction. 

Specifically, we employ 3DGS to represent each source image $\bm{I}_i$ as a set of 3D Gaussians. Each pixel in the foreground corresponds to a unique 3D Gaussian. We use $U_i$ to denote the number of Gaussians for source image $i$. The proposed EVA-Gaussian predicts the positions and attributes of 3D Gaussians in the form of attribute maps $\{\bm{M}_i\}_{i=1}^{n}=\{\bm{P}_i, \bm{O}_i, \bm{S}_i, \bm{Q}_i, \bm{F}_i\}_{i=1}^{n}$ from the image set $\{\bm{I}_i\}_{i=1}^{n}$, where $\bm{P}_i, \bm{O}_i, \bm{S}_i, \bm{Q}_i$, and $\bm{F}_i$ denote the attribute maps for Gaussian positions, opacities, scales, quaternions, and features of source image $i$, respectively. Notably, in the feature map $\bm{F}_i = \{\bm{f}_i^u\}_{u=1}^{U_i}$, each element $\bm{f}_i^u \in \mathbb{R}^{32}$ serves as a new attribute associated with each 3D Gaussian, which will be used later in Sec. \ref{section Feature Splatting and Refinement} to remove artifacts caused by geometric errors in $\{\bm{P}_i\}_{i=1}^{n}$. Mathematically, the procedure of EVA-Gaussian is expressed as:
 \begin{equation}
     \{\bm{M}_i\}_{i=1}^{n} = \mathcal{D}_{\bm{\theta}}(\{\bm{I}_i\}_{i=1}^{n}),
 \end{equation}
where $\bm{\theta}$ denotes the learnable parameters of the network.

The framework of EVA-Gaussian is depicted in Fig. \ref{fig:network_overview}. EVA-Gaussian splits the process of predicting Gaussian maps into three stages. In the first stage,
it employs a U-Net architecture with an Efficient cross-View Attention module (EVA) to obtain enhanced multi-view predictions of the 3D Gaussian position maps $\{\bm{P}_i\}_{i=1}^{n}$, as elaborated in Sec. \ref{Gaussian Position Estimation}. In the second stage, a Gaussian attribute prediction network, detailed in Sec. \ref{Gaussian Attributes Estimation}, takes the predicted position maps $\{\bm{P}_i\}_{i=1}^{n}$ and the original RGB images $\{\bm{I}_i\}_{i=1}^{n}$ as input to estimate the remaining attributes of 3D Gaussians. The predicted 3D Gaussians from all source images are then aggregated to render target views using differential rasterization \citep{3dgs}. In the final stage, the rendered image $\hat{\bm{I}}^0$ and its corresponding feature map $\bm{F}_{\text{novel}}$ are fused for further refinement using the network described in Sec. \ref{section Feature Splatting and Refinement}. In addition, an anchor loss is introduced during the training stage to enhance the overall reconstruction quality, as depicted in Sec. \ref{Regularization}.

\begin{figure*}[t]
    \centering
    \includegraphics[width=0.9\textwidth]{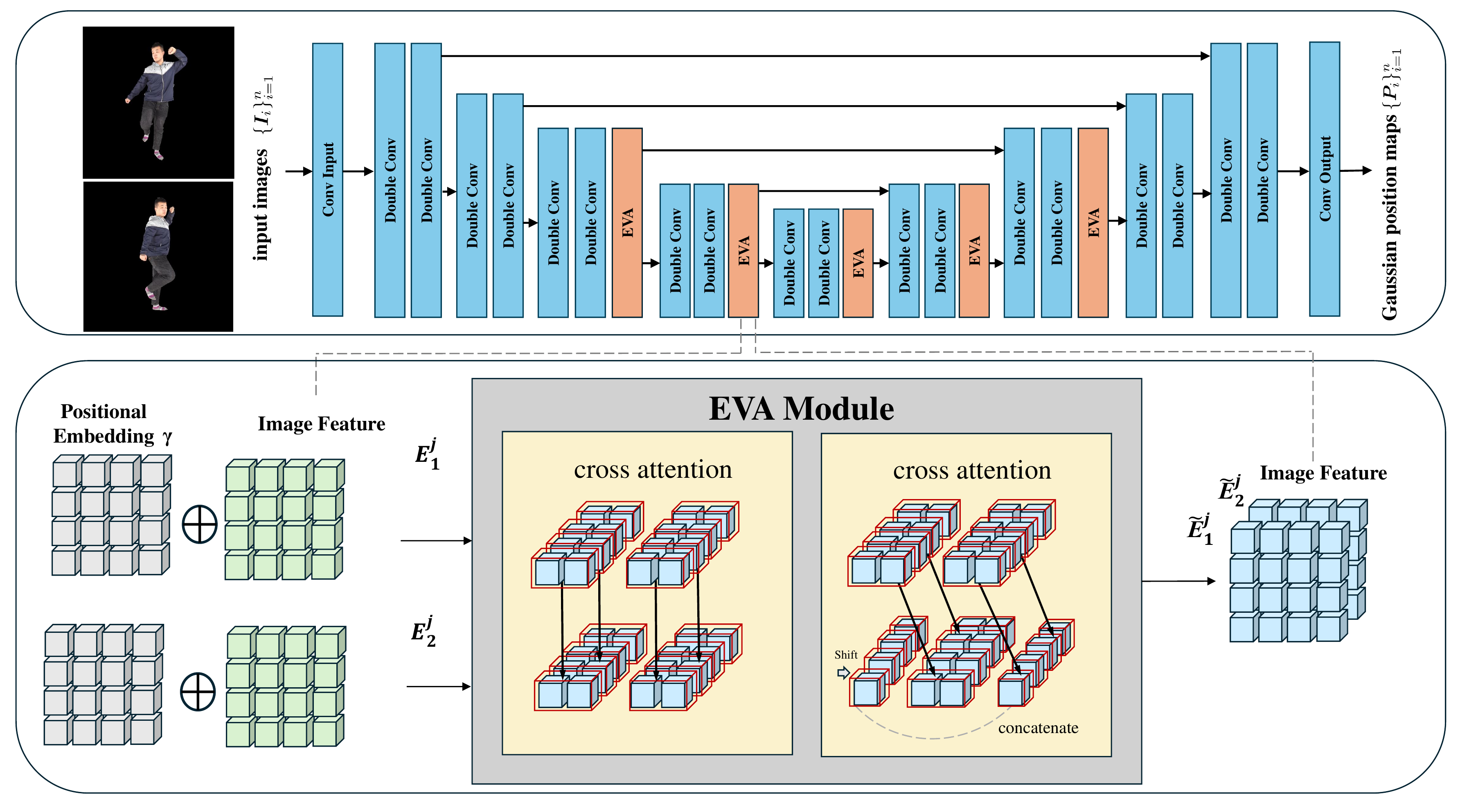}
    \vspace{-4pt}
    \caption{\textbf{Efficient cross-View Attention (EVA) module} for 3D Gaussian position estimation. EVA takes multi-view image features as input, embeds them into window patches using a shifted algorithm, and performs cross attention between the features from different views.}
    \label{fig:EVA_module}
    \vspace{-4pt}
\end{figure*}

\subsection{Gaussian Position Estimation}
\label{Gaussian Position Estimation}
The variations in depth across the surface of human body may appear minimal. However, these nuances are critically important, particularly in regions such as the face and hands that contain a wealth of semantic information. Even slight inaccuracies in depth estimation within these areas can lead to significant degradation in visual quality and fidelity. This underscores the necessity for precise estimation of 3D Gaussian positions to enable effective and high-fidelity human reconstruction. 

To tackle this challenge, we employ a U-Net based architecture, denoted by $\mathcal{D}^{P}_{\bm{\theta}_1}$, to estimate the 3D Gaussian position maps $\{\bm{P}_i\}_{i=1}^{n}$ from multi-view images $\{\bm{I}_i\}_{i=1}^{n}$. This process is expressed as:
\begin{equation}
    \{\bm{P}_i\}_{i=1}^{n} = \mathcal{D}^{P}_{\bm{\theta}_1}(\{\bm{I}_i\}_{i=1}^{n}).
\end{equation}
To ensure accurate depth estimation across diverse camera angles and arbitrary input views at high resolutions, while maintaining computational and temporal efficiency, we propose an EVA module, as illustrated in Fig. \ref{fig:EVA_module}. This module is integrated into the three lowest resolution layers of the U-Net basebone $\mathcal{D}^{P}_{\bm{\theta}_1}$ to facilitate the multi-view correspondence retrieval and information exchange. We use $j$ to index these three layers, with $j = -1,j = -2,$ and $j = -3$ representing the lowest, second-lowest, third-lowest resolution layers, respectively.
The EVA module takes intermediate image features $\boldsymbol{E}^j_i \in \mathbb{R}^{R^j\times C^j}, \forall i \in \{1, \cdots, n\}, \forall j \in \{-1, -2, -3\}$, as input, where $R^j$ and $C^j$ denote the total number of pixels and the channel dimension of each pixel at layer $j$, respectively. The module outputs enhanced image features $\widetilde{\boldsymbol{E}}^j_i$, which incorporate cross-view information. Before the execution of attention mechanisms, a learnable positional embedding $\gamma$ is added to encode spatial coordinates, improving the understanding of image geometry.

While cross-view attention mechanisms have been explored in various vision tasks that require multiple image inputs, existing approaches in feed-forward 3D Gaussian reconstruction, such as epipolar attention in pixelSplat \citep{pixelsplat} and vanilla cross-attention in MVSplat \citep{mvsplat}, are limited to low-resolution inputs (256$\times$256) and fail to recover fine-grained details. In contrast, our approach targets high-resolution (1024$\times$1024) human-centric reconstruction, a setting that has been demonstrated to be critical for recovering intricate human details \citep{pifuhd}. The key novelty of our EVA module lies in its localized 1D window-based attention mechanism, which exploits the observation that corresponding pixels from reference views are typically located in adjacent positions along the x-axis in human-centric camera settings. Unlike global attention mechanisms used in prior works \citep{mvsplat, pixelsplat, FGGS}, EVA computes cross-attention only within a local 1D window aligned with the x-axis, and the window is shifted by half its size at each iteration to expand the receptive field. This design significantly reduces computational complexity and improves GPU memory utilization, while maintaining high performance, as demonstrated in Tab. \ref{attention}.


Notably, when the scale of each Gaussian is sufficiently small, the 3D Gaussian position of a pixel aligns precisely with its corresponding value on the depth map. A detailed proof of this property is provided in Appendix D. Based on this observation, we train the position estimation network $\mathcal{D}^{P}_{\bm{\theta}}$ to obtain the position maps $\{\bm{P}_i\}_{i=1}^{n}$ with the mean squared error (MSE) loss function:
\begin{equation}
    \mathcal{L}_{\text{depth}} = ||\bm{P}_i-\bm{P}_i^{\text{gt}}||_2,
\end{equation}
where $\bm{P}_i^{\text{gt}}$ denotes the ground truth depth map.

\subsection{Gaussian Attribute Estimation}
\label{Gaussian Attributes Estimation}
To complete the estimation of 3D Gaussian maps $\{\bm{M}_i\}_{i=1}^{n}$, we employ a shallow U-Net $\mathcal{D}^{A}_{\bm{\theta}_2}$ to estimate the remaining attributes $\bm{O}_i$, $\bm{S}_i$, $\bm{Q}_i$, $\bm{F}_i$. This network takes the estimated position maps $\{\bm{P}_i\}_{i=1}^{n}$ from the first stage in Sec. \ref{Gaussian Position Estimation} and the original RGB images $\{\bm{I}_i\}_{i=1}^{n}$ as input, and outputs 3D Gaussian attributes $\bm{O}_i$, $\bm{S}_i$, $\bm{Q}_i$, $\bm{F}_i$, which is expressed as: 
\begin{equation}
    \{\bm{O}_i, \bm{S}_i, \bm{Q}_i, \bm{F}_i\}_{i=1}^{n}=\mathcal{D}^{A}_{\bm{\theta}_2}(\{\bm{I}_i\}_{i=1}^{n}\oplus\{\bm{P}_i\}_{i=1}^{n}).
\end{equation}
The resulting estimated 3D Gaussian maps $\{\bm{M}_i\}_{i=1}^{n}=\{\bm{P}_i, \bm{O}_i, \bm{S}_i, \bm{Q}_i, \bm{F}_i\}_{i=1}^{n}$ are then utilized to render novel views through the $\alpha$-blending mechanism, as in the vanilla 3DGS \citep{3dgs}.
The network $\mathcal{D}^{A}_{\bm{\theta}_2}$ is trained by using a combination of MSE loss and structural similarity index measure (SSIM) \citep{ssim} loss between the rendered novel view image $\hat{\bm{I}}^0$ and the ground truth $\bm{I}^{\text{gt}}$ as follows:
\begin{equation}
  \mathcal{L}_{\text{render}}= ||\hat{\bm{I}}^0-\bm{I}^{\text{gt}}||_2 +\lambda_{\text{render}}(1-\text{SSIM}(\hat{\bm{I}}^0, \bm{I}^{\text{gt}})),  \label{render_image_loss}
\end{equation}
where $\lambda_{\text{render}}$ denotes the weighting factor for the SSIM loss.

\subsection{Feature Splatting and Refinement}
\label{section Feature Splatting and Refinement}
The 3D Gaussian position maps $\bm{P}_i$ estimated in Sec. \ref{Gaussian Position Estimation} inevitably contain some degree of error, which may lead to distortions and artifacts in the rendered RGB images. To mitigate these issues, we propose a post-splatting refinement method to correct the position estimates. Recent studies \citep{featuresplat} have demonstrated that feature vector representations can capture scene information more effectively than spherical harmonics, resulting in significant improvements in novel view synthesis, particularly in scenarios with limited overlapping views. Inspired by this finding, we incorporate a feature vector, i.e., $\bm{f}^u_i \in \mathbb{R}^{32}$ mentioned in Sec. \ref{section Overview}, as an additional attribute for each Gaussian to more precisely capture its spatial characteristics.

During the splatting process, we first aggregate the 3D Gaussians from all source views. Then, the color values of these 3D Gaussians are rendered using $\alpha$-blending mechanism from 3D Gaussian Splatting \citep{3dgs}. Concurrently, the feature values of the 3D Gaussians are splatted onto the image plane using a modified $\alpha$-blending function: 
\begin{equation}
    \bm{f}_{\text{pixel}} = \sum_{j=1}^{N}\bm{f}_j \alpha_{j}\prod_{l=1}^{j-1}(1-\alpha_{l}),  \label{renderfeat}
\end{equation}
where $\bm{f}_\text{pixel}$ is the feature vector for the corresponding pixel on the feature map of the novel view image $\bm{F}_\text{novel}$, $\bm{f}_j$ denotes the feature vector for the 3D Gaussian with the $j$-th greatest depth, and $N=\sum_{i=1}^{n}U_i$ is the total number of 3D Gaussians from all source views. 

Moreover, we employ a carefully designed recurrent U-Net $\mathcal{D}_{\theta_3}^{R}$ that takes both the RGB and feature images as input and project them onto the RGB space for the final output through $L$ recurrent loops. This procedure is expressed as:
\begin{align}
 &\hat{\bm{I}}^{l} = \mathcal{D}_{\bm{\theta}_3}^{R}(\hat{\bm{I}}^{l-1}\oplus \bm{F}_{\text{novel}}), l \in \left \{ 1 \cdots L \right \}, \label{recurrent_func}
\end{align}
where 
 $\hat{\bm{I}}^{l} \in \mathbb{R}^{H\times W \times 3}$ and $\bm{F}_{\text{novel}} \in \mathbb{R}^{H\times W \times 32}$. Similar to the Gaussian attribute estimation, the loss function for supervising the final output is a combination of the MSE loss and the SSIM loss between the refined image $\hat{\bm{I}}^L$ and the ground truth $\bm{I}^{\text{gt}}$ as follows:
\begin{equation}
  \mathcal{L}_{\text{refine}}= ||\hat{\bm{I}}^L-\bm{I}^{\text{gt}}||_2 +\lambda_{\text{refine}}(1-\text{SSIM}(\hat{\bm{I}}^L, \bm{I}^{\text{gt}})),   \label{refine_image_loss}
\end{equation}
where $\lambda_{\text{refine}}$ denotes the weighting factor for the SSIM loss.

\begin{table}[t]
\caption{Comparison of computational and temporal costs for different attention mechanisms. CVA refers to the cross-view attention in MVSplat, EA denotes the epipolar attention in pixelSplat, and w16, w32, w64 represent the window sizes of 16, 32, and 64, respectively, in EVA. G.M. indicates the GPU memory usage.
}
\label{attention}
\resizebox{1.\linewidth}{!}{
\begin{tabular}{ccc|cc|cc}
\hline
Input Size & \multicolumn{2}{c|}{2$\times$64$\times$128$\times$128} & \multicolumn{2}{c|}{2$\times$64$\times$256$\times$256} & \multicolumn{2}{c}{2$\times$32$\times$256$\times$256} \\ \hline
Module & Time  & G.M. & Time & G.M. & Time & G.M. \\ \hline
CVA & 35.3ms & 3.72GB & 30.4ms & 35.44GB & 26.3ms & 31.77GB \\ \hline
EA & 58.3ms & 15.19GB & 19.3ms & 59.14GB & 16.9ms & 58.01GB \\ \hline
EVA (w16) & 7.23ms & 0.92GB & 1.77ms & 2.14GB & 1.43ms & 1.37GB \\ \hline
EVA (w32) & 6.53ms & 0.92GB & 1.49ms & 2.14GB & 1.16ms & 1.37GB \\ \hline
EVA (w64) & 6.31ms & 0.92GB & 1.39ms & 2.14GB & 1.06ms & 1.37GB \\ \hline
\end{tabular}}
\vspace{-10pt}
\end{table}

\subsection{Attribute Regularization}
\label{Regularization}
Existing feed-forward 3D Gaussian Splatting methods, such as GPS-Gaussian \cite{gpsgs} and MVSplat \cite{mvsplat}, typically assume that the position of 3D Gaussians correspond directly to the estimated depth values in their respective views. However, this assumption does not always hold, as 3D Gaussians are not explicitly constrained to lie on the surface of objects. Moreover, the unprojection process in these methods is performed independently for each view using an orthogonal projection framework. This approach lacks explicit scene understanding and cross-view consistency, as it does not incorporate mechanisms for cross-view matching or global scene optimization during the standard training procedure.

To address these limitations and improve reconstruction quality, we introduce a regularization term to enhance the training process. Specifically, we propose an anchor loss to regularize the scales and opacities of Gaussians, ensuring consistency between the geometry of predicted depth maps and the 3D Gaussian positions, as proved in Appendix D. This loss also aligns the Gaussians from different views to force their locations to the same landmark. We leverage MediaPipe \citep{mediapipe} to annotate human facial landmarks and compute the anchor loss to regularize the 3D landmark Gaussian scales, opacities, and positions as follows:
\begin{align}
\mathcal{L}_{\text{anchor}}=&\lambda_{\text{opacity}}\sum_{i=1}^{N} ||\bm{O}_i\text{log}(\bm{O}_i)||_1 +  \lambda_{\text{scale}}\sum_{i=1}^{N} ||\bm{S}_i||_2+  \nonumber\\
   &\sum_{i,j\in \mathbb{V}}\sum_{m_i\in \mathbb{M}_i, m_j\in \mathbb{M}_j}  \max\left \{ Dist, t\right \},   \label{anchor}
\end{align}
where $Dist$ represents the distance of human facial landmarks in 3D space with 
\begin{align*}
    Dist = ||\Pi^{-1}(\bm{m}_i, \bm{P}_i(\bm{m}_i)) - \Pi^{-1}(\bm{m}_j, \bm{P}_j(\bm{m}_j))||_2,
\end{align*} 
$\{\mathbb{M}_i\}_{i=1}^{n}$ denotes the collection of all landmarks on the 2D image plane, $\mathbb{V}$ denotes the collection of source views, and $\Pi^{-1}$ represents the process of reprojection from 2D image to 3D space. In addition, a tolerance factor $t$ is introduced to mitigate the errors in the estimated landmarks.


By integrating the loss functions from the three stages, i.e., $\mathcal{L}_{\text{depth}}, \mathcal{L}_{\text{render}}, \mathcal{L}_{\text{refine}}$, and the proposed regularization term $\mathcal{L}_{\text{anchor}}$, the overall training loss for EVA-Gaussian is:
\begin{equation}
   \mathcal{L}_{\text{EVA}}= \mathcal{L}_{\text{depth}}+\lambda_1 \mathcal{L}_{\text{render}}+ \lambda_2 \mathcal{L}_{\text{refine}} +\lambda_3 \mathcal{L}_{\text{anchor}}, \label{lossfunc}
\end{equation}
where $\lambda_1$, $\lambda_2$, and $\lambda_3$ are weighting factors.

Since the 3D Gaussian position and attribute estimation stages can be executed within tens of milliseconds, and the feature refinement stage is lightweight, taking less than ten milliseconds, EVA-Gaussian is capable of rapidly reconstructing 3D human subjects from a collection of RGB images and rendering novel views in a real-time manner.

\section{Experiments}

\begin{table}[t]
\caption{Comparison with feed-forward 3D reconstruction methods at a resolution of 256$\times$256. Inference speeds are reported in the last column.
Better results are marked in a deeper color.
}
\vspace{-8pt}
\label{stereoscene}
\begin{center}
\resizebox{\linewidth}{!}{
\begin{tabular}{cccccccc}
\hline
 \multirow{2}{*}{$\Delta$=45°} & \multicolumn{3}{c}{THuman2.0} & \multicolumn{3}{c}{THumansit} & \multirow{2}{*}{Speed}\\ \cline{2-7}
 & PSNR$\uparrow$  & SSIM$\uparrow$  & LPIPS$\downarrow$   & PSNR$\uparrow$  & SSIM$\uparrow$  & LPIPS$\downarrow$   & \\ \hline
pixelSplat & 25.19 & 0.9156 & 0.0824 & 23.31 & 0.8880 & 0.0954 & 185ms \\
MVSplat & 28.05 & 0.9515 & 0.0346 & 24.97 & 0.9223 & 0.0532  & \cellcolor{b3}70ms\\
MVSGaussian & 26.44 & \cellcolor{b3}0.9706 & 0.0283 & 25.20 & \cellcolor{b3}0.9641 &  \cellcolor{b3}0.0297 & 71ms\\
ENeRF & \cellcolor{b3}29.62 & 0.9696 &\cellcolor{b3} 0.0238 & \cellcolor{b3}27.06 & 0.9567 & 0.0334 &136ms\\
GPS-Gaussian & \cellcolor{b2}30.30 & \cellcolor{b2}0.9762 & \cellcolor{b2}0.0224 & \cellcolor{b2}28.02 & \cellcolor{b2}0.9671 & \cellcolor{b2}0.0251 & \cellcolor{b1}40ms\\
EVA-Gaussian & \cellcolor{b1}31.11 & \cellcolor{b1}0.9782 & \cellcolor{b1}0.0198 & \cellcolor{b1}29.16 & \cellcolor{b1}0.9696 & \cellcolor{b1}0.0249  & \cellcolor{b2}55ms\\ \hline
\end{tabular}}
\end{center}
\vspace{-20pt}
\end{table}

\begin{table*}[t]
\caption{Comparison of feed-forward methods under different camera angle settings at a resolution of $1024\times 1024$. Better results are marked in a deeper color. GPS-Gaussian fails to work effectively when $\Delta=90^{\circ}$, as it is unable to meet its rectification requirement.}
\vspace{-8pt}
\label{stereohuman}
\begin{center}
\resizebox{0.9\linewidth}{!}{
\begin{tabular}{ccccccccccccc}
\toprule[1.55pt]
THuman2.0&\multicolumn{3}{c}{$\Delta= 45^{\circ}$ }&\multicolumn{3}{c}{$\Delta= 60^{\circ}$}&\multicolumn{3}{c}{$\Delta= 72^{\circ}$}&\multicolumn{3}{c}{$\Delta= 90^{\circ}$}\\ 
$1024\times1024$& PSNR$\uparrow$  & SSIM$\uparrow$  & LPIPS$\downarrow$   & PSNR$\uparrow$  & SSIM$\uparrow$  & LPIPS$\downarrow$  & PSNR$\uparrow$  & SSIM$\uparrow$  & LPIPS$\downarrow$   & PSNR$\uparrow$  & SSIM$\uparrow$  & LPIPS$\downarrow$   \\\hline
ENeRF & \cellcolor{b3}27.94 & \cellcolor{b3}0.9573 & \cellcolor{b3}0.0367&
 \cellcolor{b3}26.16 & \cellcolor{b3}0.9452 & \cellcolor{b3}0.0516&\cellcolor{b2}24.61 & \cellcolor{b3}0.9309 & \cellcolor{b3}0.0705&
 \cellcolor{b2}{22.85} & \cellcolor{b2}{0.8990} & \cellcolor{b2}{0.1147}  \\
GPS-Gaussian & \cellcolor{b2}{29.63} & \cellcolor{b2}{0.9703 }& \cellcolor{b1}{0.0174}& \cellcolor{b2}{27.36} & \cellcolor{b2}{0.9630} & \cellcolor{b2}{0.0249}&
 \cellcolor{b3}{24.25} & \cellcolor{b2}{0.9519} & \cellcolor{b2}{0.0480} & / & / & /  \\
EVA-Gaussian & \cellcolor{b1}30.46 & \cellcolor{b1}{0.9730} & \cellcolor{b2}{0.0178}
 & \cellcolor{b1}{28.29} & \cellcolor{b1}{0.9654} & \cellcolor{b1}{0.0248}&\cellcolor{b1}{27.54} & \cellcolor{b1}{0.9614} & \cellcolor{b1}{0.0297} & \cellcolor{b1}{26.31} & \cellcolor{b1}{0.9555} & \cellcolor{b1}{0.0391}\\ \midrule[1.55pt]
THumansit&\multicolumn{3}{c}{$\Delta$ = 45°}&\multicolumn{3}{c}{$\Delta$ = 60°}&\multicolumn{3}{c}{$\Delta$ = 72°}&\multicolumn{3}{c}{$\Delta$ = 90°}\\ 
$1024\times1024$& PSNR$\uparrow$  & SSIM$\uparrow$  & LPIPS$\downarrow$ &PSNR$\uparrow$  & SSIM$\uparrow$  & LPIPS$\downarrow$
& PSNR$\uparrow$  & SSIM$\uparrow$  & LPIPS$\downarrow$ & PSNR$\uparrow$  & SSIM$\uparrow$  & LPIPS$\downarrow$ \\\hline
ENeRF  & \cellcolor{b3}25.61 & \cellcolor{b3}0.9397 & \cellcolor{b3}0.0494&  \cellcolor{b3}23.80 & \cellcolor{b3}0.9168 & \cellcolor{b3}0.0745 & \cellcolor{b2}22.48 & \cellcolor{b3}0.8956 & \cellcolor{b3}0.0985  & \cellcolor{b2}{21.20} & \cellcolor{b2}{0.8571} & \cellcolor{b2}{0.1406} \\
GPS-Gaussian& \cellcolor{b2}{27.05} & \cellcolor{b2}{0.9584} & \cellcolor{b1}{0.0227} & \cellcolor{b2}{25.19} & \cellcolor{b2}{0.9480}& \cellcolor{b2}{0.0351}& \cellcolor{b3}{21.48} & \cellcolor{b2}{0.9276} & \cellcolor{b2}{0.0713} & / & / & / \\
EVA-Gaussian & \cellcolor{b1}{28.76} & \cellcolor{b1}{0.9621} & \cellcolor{b2}{0.0236}& \cellcolor{b1}{27.38} & \cellcolor{b1}{0.9543} & \cellcolor{b1}{0.0321}  & \cellcolor{b1}{26.60} & \cellcolor{b1}{0.9498} & \cellcolor{b1}{0.0500}  & \cellcolor{b1}{25.44} & \cellcolor{b1}{0.9416} & \cellcolor{b1}{0.0512}  \\ \bottomrule[1.55pt]
\vspace{-32pt}
\end{tabular}
}
\end{center}
\end{table*}
\begin{figure*}[t]
    \centering
    \includegraphics[width=0.8\textwidth]{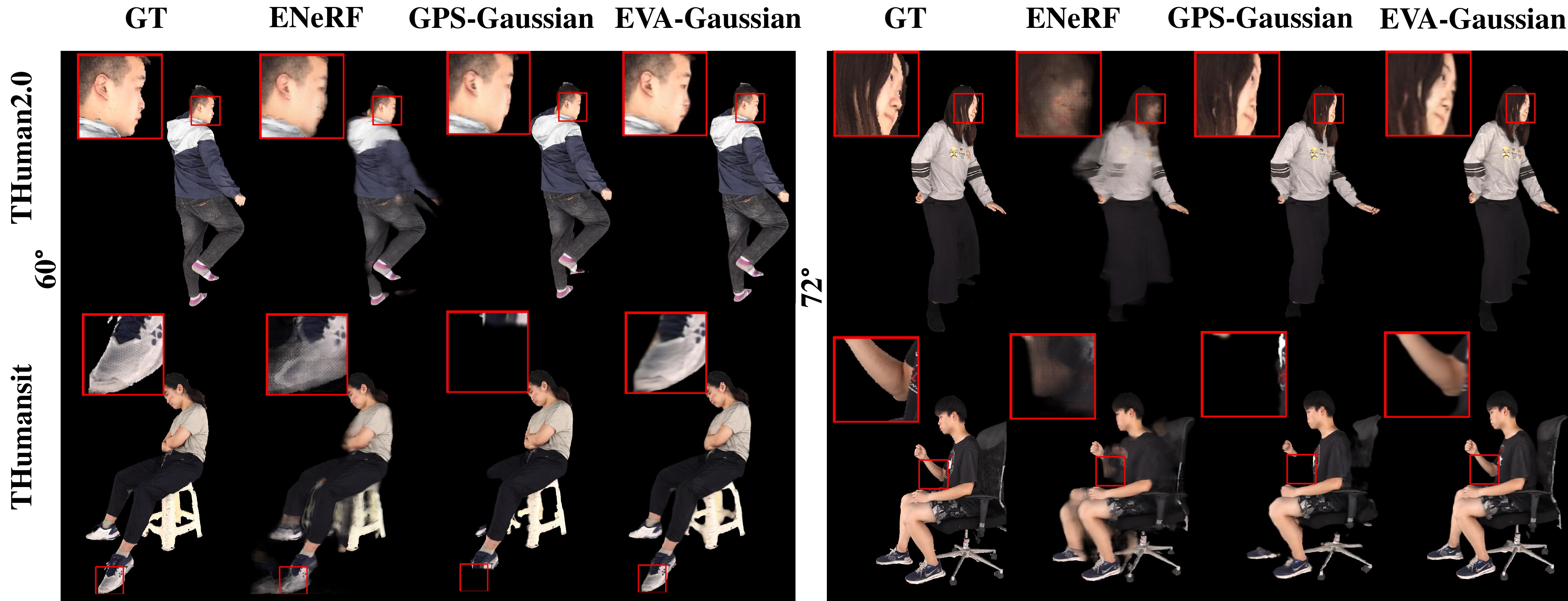}
    \caption{Qualitative comparison on THuman2.0 and THumansit. EVA-Gaussian achieves superior novel view rendering quality under diverse camera settings. Additional visualization results are provided in Appendix C.}
    \label{fig:main_vis}
\end{figure*}

\subsection{Experiment Setup}
\textbf{Implementation details.} Our EVA-Gaussian is trained on 1024$\times$1024 pixel images across multiple training views using a single NVIDIA A800 GPU for 100K iterations with the AdamW \citep{adamw} optimizer, unless otherwise specified. For the 3D Gaussian position estimation stage, it is first pretrained under the supervision of ground truth depth maps. Baselines are trained using their publicly available code. More implementation details are provided in Appendix A.

\noindent \textbf{Datasets.} We conduct experiments on two open-source human body datasets: THuman2.0 \citep{thuman2.0} and THumanSit \citep{thumansit}. THuman2.0 contains 526 unique human models with their corresponding SMPL parameters, among which 100 individuals are randomly selected for our evaluation. The THumanSit dataset has a similar structure, containing 72 human models with around 60 poses for each, and we randomly choose 5 individuals with all poses for our evaluation.

\noindent \textbf{Metrics.} We report results on commonly used metrics: PSNR, SSIM \citep{ssim}, and LPIPS \citep{lpips}, computed over the entire image, as some methods may produce artifacts outside the human bounding box \citep{enerf,gpsgs}. We also include the inference time to demonstrate the real-time reconstruction capability of our method. 

\begin{table*}[t]
\caption{Comparison with GPS-Gaussian under different camera numbers. Results in bold represent the best performance. EVA-Gaussian achieves SOTA performance across various metrics, primarily due to the multi-view consistency enabled by our proposed EVA module.}
\vspace{-8pt}
\label{multiviewhuman}
\begin{center}
\resizebox{0.9\linewidth}{!}{
\begin{tabular}{ccccccccccccc}
\hline
\multirow{2}{*}{1024$\times$1024} & \multicolumn{6}{c}{THuman2.0~($\Delta= 45^{\circ}$)}& \multicolumn{6}{c}{THumansit~($\Delta= 45^{\circ}$)} \\ \cline{2-13}
 & \multicolumn{3}{c}{3 views}& \multicolumn{3}{c}{4 views}& \multicolumn{3}{c}{3 views}& \multicolumn{3}{c}{4 views} \\ \hline
  & PSNR$\uparrow$  & SSIM$\uparrow$  & LPIPS$\downarrow$   & PSNR$\uparrow$  & SSIM$\uparrow$  & LPIPS$\downarrow$ & PSNR$\uparrow$  & SSIM$\uparrow$  & LPIPS$\downarrow$   & PSNR$\uparrow$  & SSIM$\uparrow$  & LPIPS$\downarrow$   \\ 
GPS-Gaussian & 28.74 & 0.9655 & 0.0200& 28.51 & 0.9636 & 0.0218 & 26.87& 0.9523& \textbf{0.0243}  & 26.50& 0.9498& 0.0267\\
EVA-Gaussian & \textbf{30.76} & \textbf{0.9722} & \textbf{0.0175}& \textbf{30.35} & \textbf{0.9707} & \textbf{0.0189} & \textbf{28.64} &\textbf{0.9596}  & 0.0255  & \textbf{28.32} & \textbf{0.9582} & \textbf{0.0260} \\ \hline
\end{tabular}
}
\end{center}
\vspace{-10pt}
\end{table*}

\subsection{Stereo Reconstruction}
\label{stereoreconstruction}

\textbf{Comparison with  state-of-the-art feed-forward reconstruction methods.} 
We first compare our approach against state-of-the-art (SOTA) feed-forward reconstruction methods, including ENeRF \citep{enerf}, pixelSplat \citep{pixelsplat}, MVSplat \citep{mvsplat}, MVSGaussian \citep{mvsgaussian}, and GPS-Gaussian \citep{gpsgs}.
All experiments are conducted in a stereo-view setting, where the angle between the two camera views $\Delta=45^{\circ}$. 
The attention modules in the scene reconstruction methods \citep{pixelsplat, mvsgaussian, mvsplat} are inefficient in their utilization of GPU memory, limiting their ability to train effectively at a high resolution of 1024$\times$1024. Therefore, we also conduct a fair comparison of all methods at a resolution of 256$\times$256.
The quantitative results presented in Table \ref{stereoscene} demonstrate that EVA-Gaussian achieves the best novel view quality in terms of PSNR, SSIM, and LPIPS, while maintaining the second-fastest inference speed.

\noindent\textbf{Comparison under diverse angle changes between camera views.} We further evaluate the performance of our method across four different angles between the two camera views, i.e., $\Delta=45^{\circ}, 60^{\circ}, 72^{\circ}$, and $90^{\circ}$, at a high resolution of 1024$\times$1024. As shown in Table \ref{stereohuman}, our EVA-Gaussian outperforms all baseline methods on all metrics, achieving a maximum PSNR advantage of 5.12 dB. Notably, thanks to our EVA module, EVA-Gaussian remains effective even under extremely sparse camera settings, e.g., $\Delta=90^{\circ}$. In contrast, GPS-Gaussian fails to work effectively due to its reliance on stereo rectification. Fig. \ref{fig:main_vis} presents the qualitative results of novel view rendering, where EVA-Gaussian outperforms previous SOTA methods in rendering quality, especially in scenarios with large viewpoint discrepancies.

\subsection{Multi-view Reconstruction}
We conduct experiments under multi-view settings to evaluate EVA-Gaussian's capability ti handle more than two input images. Table \ref{multiviewhuman} presents the quantitative results, where our method demonstrates a performance gain with more than 1.5 dB improvement over the baseline. While the performance of GPS-Gaussian drops significantly due to the mismatch among multiple inferences, our method maintains high performance, thanks to the cross-view consistency ensured by our proposed EVA module.

\subsection{Generalizability Validation}
\begin{figure}[t]
    \centering
    \includegraphics[width=0.48\textwidth]{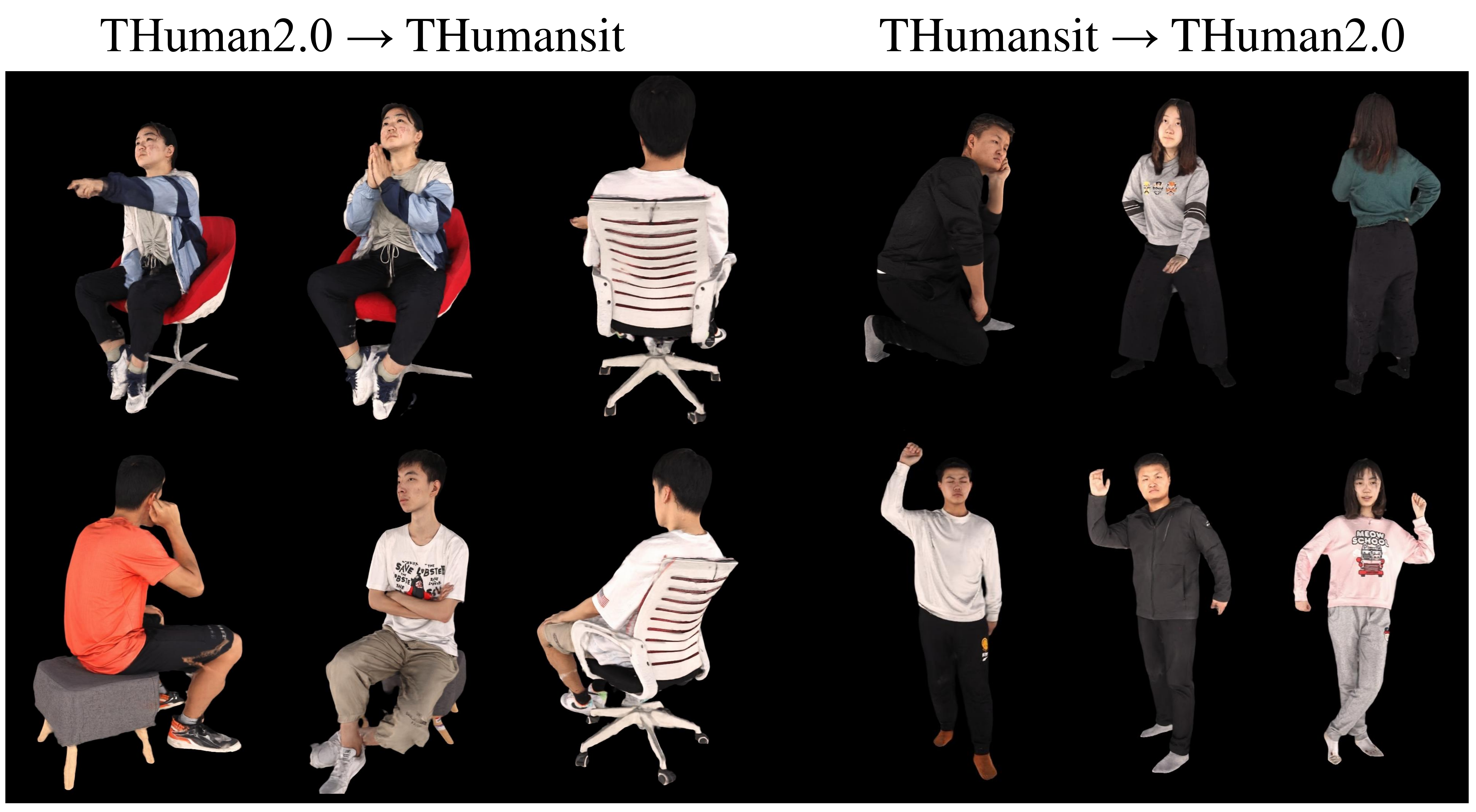}
    \caption{Visualization of cross-domain evaluation results for EVA-Gaussian. The left side displays the rendered results generated by EVA-Gaussian trained on the THuman2.0 dataset and evaluated on the THumansit dataset, while the right side shows the rendered results from EVA-Gaussian trained on the THumansit dataset and evaluated on the THuman2.0 dataset.}
    \label{fig:vis_cd}
\end{figure}

We validate the generalizability of EVA-Gaussian by taking out-of-domain images as input.
It is observed that EVA-Gaussian is inherently superior in generalizing to out-of-distribution human identities and postures, primarily due to the strong data processing ability of its attention modules in EVA-Gaussian. This allows EVA-Gaussian to maintain consistent performance when provided with sufficient data. 
As shown in Table \ref{crossdomain}, given that THumansit contains significantly more human models (over 4,000) than THuman2.0 (526 models), EVA-Gaussian shows a greater performance improvement (+2.11 dB in PSNR) compared to GPS-Gaussian (+1.97 dB in PSNR). 
This conclusion is further supported by the evaluation on THumansit, where models trained on THuman2.0 experience a performance decline due to limited data availability. Despite this, EVA-Gaussian still outperforms GPS-Gaussian, achieving a performance gain of 0.41 dB in PSNR. In addition, Fig. \ref{fig:vis_cd} provides visual evidence of EVA-Gaussian's robustness to out-of-domain data, since we have explicitly introduced inductive bias in the EVA module mitigates cross-domain performance degradation. Additional validation results on a real-world dataset are provided in Appendix B.

\begin{table}[t]
\caption{Quantitative results of cross-domain validation. EVA-Gaussian consistently outperforms GPS-Gaussian.}
\vspace{-9pt}
\label{crossdomain}
\begin{center}\resizebox{\linewidth}{!}{
\begin{tabular}{cccccccccc}
\hline
 \multirow{2}{*}{Method}& \multicolumn{3}{c}{THumansit $\rightarrow$ THuman2.0}& \multicolumn{3}{c}{THuman2.0 $\rightarrow$ THumansit}  \\ \cline{2-7}
& PSNR$\uparrow$  & SSIM$\uparrow$   & LPIPS$\downarrow$    & PSNR$\uparrow$  & SSIM$\uparrow$   & LPIPS$\downarrow$ \\ \hline
GPS-Gaussian & 29.33 & 0.9733 & 0.0325& 20.86 & 0.9243 & 0.0872 \\
EVA-Gaussian & 30.40 & 0.9751 & 0.0321& 21.27 & 0.9275 & 0.0876 \\\hline
\end{tabular}}
\end{center}
\vspace{-16pt}
\end{table}

\subsection{Ablation Study}
\begin{figure}[!thp]
    \centering
    \vspace{-6pt}
    \includegraphics[width=0.48\textwidth]{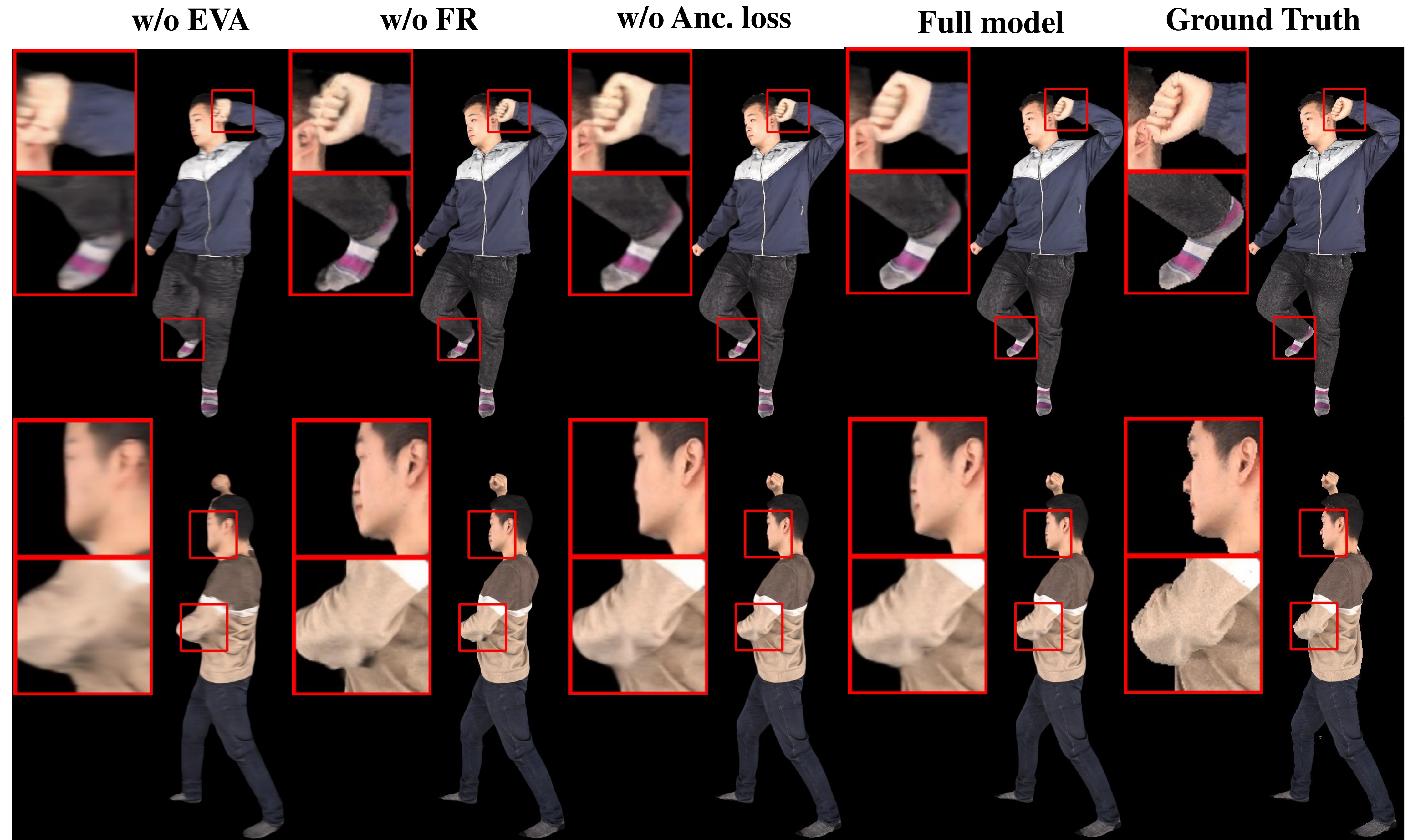}
    \caption{Qualitative visualization results of the ablation study on THuman2.0. Each module shows its effectiveness for a better visual output. The feature refinement (FR) module corrects geometric errors in the initial estimations, and the anchor loss further refines critical areas, such as the face, for generating novel view images with higher fidelity.}
    \label{fig:ablation}
    \vspace{-5pt}
\end{figure}

We conduct a detailed ablation study on THuman2.0 in a stereo-view setting, where the angle between the two views $\Delta=45^{\circ}$, as shown in Table \ref{ablation_table} and Fig. \ref{fig:ablation}. We gradually incorporate the EVA module, feature refinement module, and anchor loss to evaluate their individual contributions. Removing the EVA module results in significant degradation across all metrics, as the network struggles to perform multi-view 3D Gaussian geometry prediction. When feature refinement is excluded, artifacts appear in critical areas, such as the hands and feet. Moreover, the lack of anchor loss leads to unreliable geometry predictions, particularly in the facial region, which in turn degrades the performance across all metrics, with a notable impact on LPIPS.

\begin{table}[t]
\caption{Quantitative results of the ablation study on THuman2.0 in a stereo-view setting, where the angle between the two views $\Delta=45^{\circ}$, at a resolution of $1024\times 1024$. 
}
\vspace{-10pt}
\label{ablation_table}
\begin{center}
\resizebox{\linewidth}{!}{
\begin{tabular}{ccccc}
\hline
\multirow{2}{*}{1024×1024} & \multicolumn{4}{c}{THuman2.0~($\Delta = 45^{\circ}$)} \\ \cline{2-5} 
 &  w/o EVA & w/o f.r. &  w/o anchor loss & Full model\\ \hline
PSNR$\uparrow$ & 23.41 & 29.31 & 30.34& 30.46 \\
 SSIM$\uparrow$ & 0.9380 & 0.9676 & 0.9724 & 0.9730\\
LPIPS$\downarrow$& 0.0659 & 0.0191 & 0.0186 & 0.0178 \\ \hline
\end{tabular}}
\vspace{-10pt}
\end{center}
\end{table}

\section{Conclusion}
In this paper, we introduce EVA-Gaussian, a novel real-time 3D human reconstruction pipeline that employs multi-view attention-based 3D Gaussian position estimation and comprehensive feature refinement. To ensure robust performance, the method is trained using both photometric loss and anchor loss. Quantitative and qualitative evaluations on benchmark datasets demonstrate that EVA-Gaussian achieves state-of-the-art performance while maintaining a competitive inference speed, particularly under sparse camera settings.

While EVA-Gaussian synthesizes high-fidelity novel views, there remain several areas for improvement. For instance, the attention module can consume substantial GPU memory when processing a large number of input views or high-resolution images. In addition, the naive reprojection of pixels into 3D space may introduce conflicts in overlapping areas, leading to redundancy in the 3D representation. These limitations can be effectively addressed by incorporating RGB-D information or developing advanced techniques for detecting and resolving overlapping areas.


\newpage
\appendix

{
    \small
    \bibliographystyle{ieeenat_fullname}
    \bibliography{main}
}


\end{document}